\documentclass{article}

\usepackage[preprint]{neurips_2025}

\usepackage[utf8]{inputenc} 
\usepackage[T1]{fontenc}    
\usepackage{hyperref}       
\usepackage{xcolor}

\hypersetup{
    colorlinks=true,
    linkcolor=magenta,
    urlcolor=magenta
}
\usepackage{url}            
\usepackage{booktabs}       
\usepackage{amsfonts}       
\usepackage{nicefrac}       
\usepackage{microtype}      
\usepackage{xcolor}         
\usepackage{amsmath}
\usepackage{graphicx}
\usepackage{comment}
\usepackage{bm}
\usepackage{enumitem}

\usepackage{amsmath,amsfonts,bm}

\def\eqref#1{equation~\ref{#1}}
\def\Eqref#1{Equation~\ref{#1}}

\def\1{\bm{1}}

\DeclareMathAlphabet{\mathsfit}{\encodingdefault}{\sfdefault}{m}{sl}
\SetMathAlphabet{\mathsfit}{bold}{\encodingdefault}{\sfdefault}{bx}{n}

\title{DS-VTON: An Enhanced Dual-Scale Coarse-to-Fine Framework for Virtual Try-On}

\author{
\small 
Xianbing Sun$^{1}$, Yan Hong$^{2}$, Jiahui Zhan$^{1}$, Jun Lan$^{2}$, Huijia Zhu$^{2}$, Weiqiang Wang$^{2}$\\
\small
\textbf{Liqing Zhang}$^{1}$, \textbf{Jianfu Zhang}$^{1}$\thanks{Corresponding author.}
\\[2pt] \small 
$^{1}$Shanghai Jiao Tong University
\\[2pt] \small 
$^{2}$Ant Group 
\\[2pt] \small 
\{fufengsjtu, c.sis\}@sjtu.edu.cn \\
}

\begin{document}

\maketitle

\begin{figure}[!h]
  \centering
  \includegraphics[width=\linewidth]{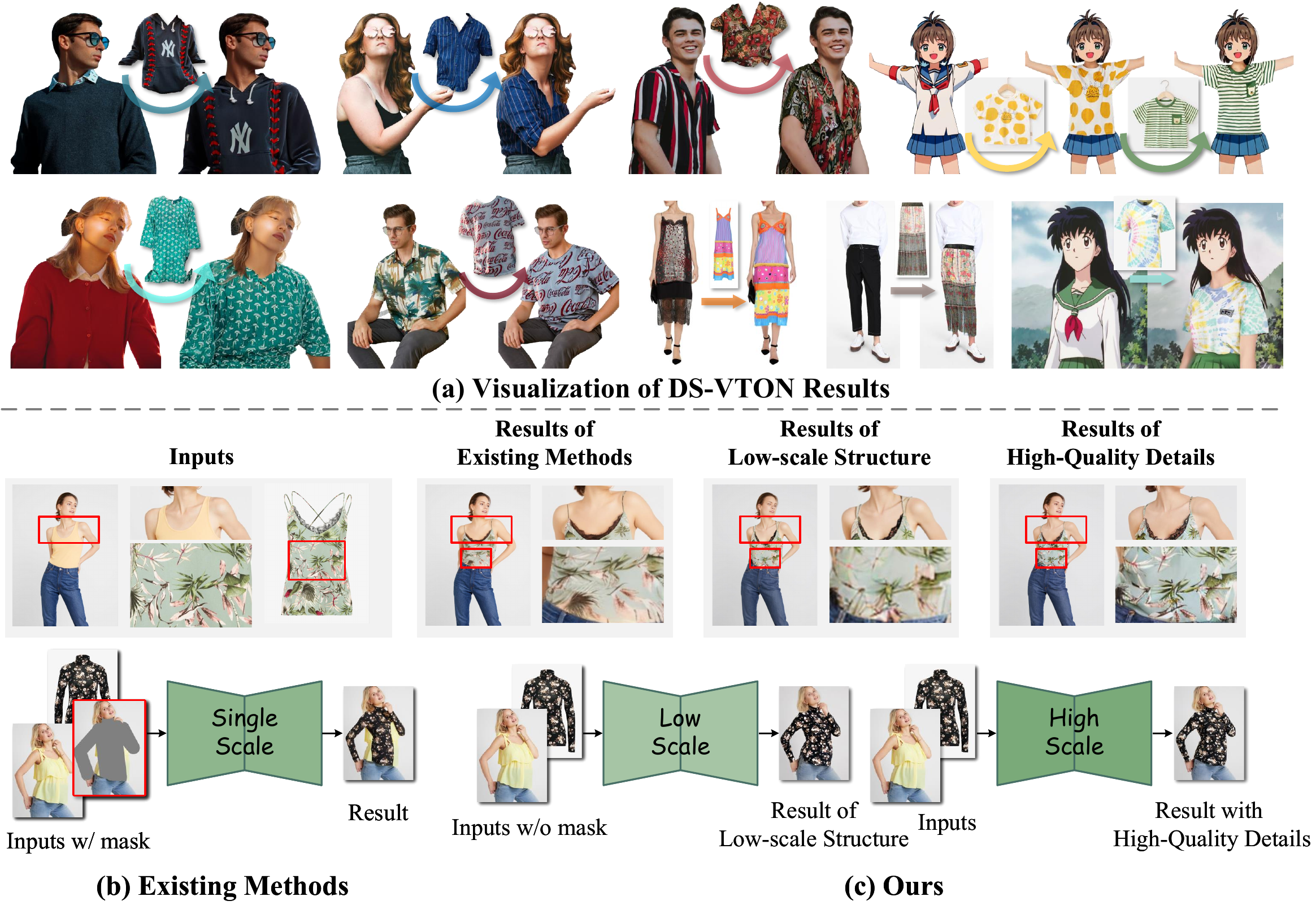}
  \caption{(a) DS-VTON results across diverse scenarios. (b) Existing methods~\citep{kim2024stableviton,xu2025ootdiffusion,choi2025improving,zhou2025flowattention} adopt a single-scale pipeline with masked inputs, limiting their ability to capture full-body semantics and garment structure. (c) In contrast, DS-VTON adopts an enhanced dual-scale coarse-to-fine framework combined with a mask-free strategy.}
  \label{fig:teaser}
\end{figure}

\begin{abstract}
 Despite recent progress, most existing virtual try-on methods still struggle to simultaneously address two core challenges: accurately aligning the garment image with the target human body, and preserving fine-grained garment textures and patterns. These two requirements map directly onto a coarse-to-fine generation paradigm, where the coarse stage handles structural alignment and the fine stage recovers rich garment details. Motivated by this observation, we propose DS-VTON, an enhanced dual-scale coarse-to-fine framework that tackles the try-on problem more effectively. DS-VTON consists of two stages: the first stage generates a low-resolution try-on result to capture the semantic correspondence between garment and body, where reduced detail facilitates robust structural alignment. In the second stage, a blend-refine diffusion process reconstructs high-resolution outputs by refining the residual between scales through noise–image blending, emphasizing texture fidelity and effectively correcting fine-detail errors from the low-resolution stage. In addition, our method adopts a fully mask-free generation strategy, eliminating reliance on human parsing maps or segmentation masks. Extensive experiments show that DS-VTON not only achieves state-of-the-art performance but consistently and significantly surpasses prior methods in both structural alignment and texture fidelity across multiple standard virtual try-on benchmarks. More visualizations are available in our \href{https://fu61.github.io/ds-vton.github.io/}{\textcolor{magenta}{project page}}.
\end{abstract}

\section{Introduction}
\label{sec:introduction}
Given a garment image and a person image, the goal of virtual try-on is to synthesize a photorealistic image of the person wearing the specified garment~\citep{han2018viton}. As a key enabling technology for online fashion and e-commerce, virtual try-on has attracted increasing attention in recent years~\citep{choi2021viton,ge2021parser,gou2023taming,lee2022high,zhang2024two,morelli2022dress,morelli2023ladi,choi2025improving,chong2025catvton,zhou2025flowattention}.

This task involves two fundamental challenges: (1) accurately fitting the garment onto the human body, and (2) preserving fine-grained garment textures. 
Existing methods fall into two main categories: Generative Adversarial Networks (GANs)~\citep{goodfellow2020generative} and Diffusion Models~\citep{ho2020denoising,rombach2022high}.
Early GAN-based approaches~\citep{choi2021viton,ge2021disentangled,lee2022high,xie2023gp} typically follow a two-stage pipeline: a warping module first aligns the garment with the target pose, followed by a generation module to synthesize the final image. While warping helps preserve garment appearance, the subsequent generation stage often leads to detail loss due to imperfect feature fusion.

Recent diffusion-based methods~\citep{kim2024stableviton,zhu2023tryondiffusion,xu2025ootdiffusion,choi2025improving,zhou2025flowattention} have become popular. The denoising process naturally progresses from coarse to fine: early steps capture global structure, while later steps refine texture details~\citep{balaji2022ediff,choi2022perception}. This progressive generation order aligns well with the requirements of virtual try-on, where both alignment and texture fidelity are essential.
\textbf{However, relying on a single-stage diffusion process remains inherently limited}. In practice, existing approaches still struggle to ensure accurate garment-body alignment and high-fidelity detail reconstruction. Without explicitly disentangling structure from detail, the unified framework often produces compromised visual quality.

To overcome the limitations of conventional single-stage denoising, we introduce a dual-scale coarse-to-fine framework that explicitly separates global structure alignment from fine-grained texture restoration. \textbf{Low-resolution stage:} In the first stage, the model generates a coarse try-on result by suppressing high-frequency content, emphasizing structural alignment through the low-resolution representation. \textbf{High-resolution stage:} In the second stage, a \textbf{blend-refine diffusion process} \textbf{explicitly} transforms the low-resolution output into high resolution, restoring fine textures and correcting fine-detail errors from the first stage.

Beside, traditional virtual try-on methods rely on human parsing masks for spatial guidance, our approach adopts a fully \textbf{mask-free strategy} that eliminates this dependency by leveraging the strong semantic priors embedded in pretrained diffusion models.
Our main contributions are as follows:
\begin{itemize}[leftmargin=*]
\item We propose a novel dual-scale, mask-free framework that enhances the coarse-to-fine process and is particularly well-suited for the try-on task.
\item We introduce a \textbf{blend-refine diffusion process} that \textbf{explicitly} bridges two complex distributions, enabling controllable transition from coarse alignment to fine-detail restoration.
\item Extensive experiments on VITON-HD~\citep{choi2021viton} and DressCode~\citep{morelli2022dress} demonstrate state-of-the-art performance, validating the effectiveness of our method both qualitatively and quantitatively.
\end{itemize}

\section{Related works}
\paragraph{GAN-based virtual try-on.}
Earlier methods~\citep{choi2021viton,ge2021disentangled,lee2022high,xie2023gp}, which are based on Generative Adversarial Networks (GANs)~\citep{goodfellow2020generative}, typically decompose the virtual try-on task into two stages: (1) warping the garment to align with the human body shape, and (2) integrating the warped garment with the human image to generate the final result. 
For instance, ACGPN~\citep{yang2020agpan} employs a warping module based on Thin-Plate Spline (TPS)~\citep{duchon1977splines} to deform the garment. PFAFN~\citep{ge2021parser} proposes a parser-free method that guides the garment warping process using learned appearance flows. VITON-HD~\citep{choi2021viton} introduces a specialized normalization layer and generator design to better handle garment-body misalignment during synthesis.

However, a key limitation of GAN-based approaches is their constrained capacity in capturing both precise spatial alignment and fine-grained garment details. As a result, these methods often rely heavily on the warping stage to encode garment appearance early in the pipeline. Yet, the subsequent integration stage frequently introduces artifacts or detail loss, degrading the overall realism of the try-on result.

\paragraph{Diffusion-based virtual try-on.}
With the rapid advancement of diffusion models~\citep{ho2020denoising,rombach2022high}, powerful virtual try-on methods have emerged~\citep{kim2024stableviton,zhu2023tryondiffusion,choi2025improving,zhou2025flowattention,sun2024outfitanyone}. Early methods like DCI-VTON~\citep{gou2023taming} use a two-stage pipeline: warping the garment to match the body, then blending it with the person image using a diffusion model. DT-VTON~\citep{zhang2024two} splits the task into structural alignment and texture replacement. More recent methods~\citep{morelli2023ladi,kim2024stableviton,zhu2023tryondiffusion,choi2025improving} employ a single diffusion process for direct try-on synthesis, with improved conditioning strategies. For instance, LaDI-VTON~\citep{morelli2023ladi} uses textual inversion to encode garment identity, while IDM-VTON~\citep{choi2025improving} introduces GarmentNet for structural and appearance guidance. Leffa~\citep{zhou2025flowattention} proposes a Leffa loss to guide attention weights during the final 500 denoising steps for better texture recovery.  
Complementary to these, FitDiT~\citep{jiang2024fitdit} adopts a DiT-based architecture~\citep{peebles2023scalable} with an aggressive rectangular mask to address alignment issues in mask-based pipelines.  

Despite progress, diffusion-based methods still face challenges, such as garment fragmentation due to imprecise segmentation and poor rendering of fine patterns like flowers or text, which remain key areas for improvement.

\begin{figure}[t]
  \centering
  \includegraphics[width=\linewidth]{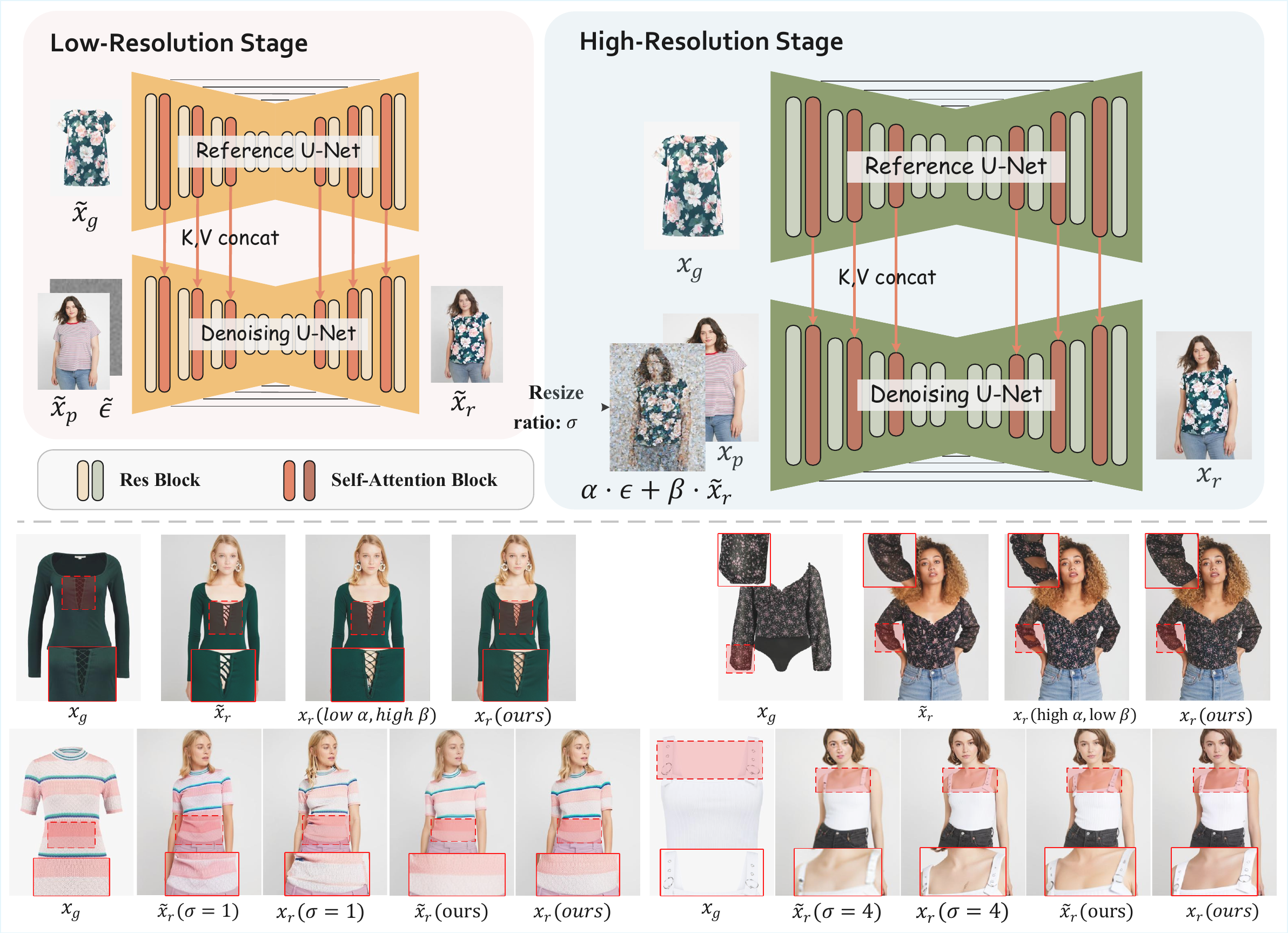}
  \caption{Upper panel: Two-scale generation pipeline. A low-resolution stage produces a coarse try-on result, then refined by a high-resolution stage; both stages share the same network architecture (see Section \ref{sec:method}). Lower panel: Results with different settings; ours uses \(\sigma = 2\) and \(\alpha = \beta = \tfrac{1}{2}\)(see Subsections \ref{subsec:low_res} and \ref{subsec:high_res}).
  With proper two-stage settings, the second stage leverages the reliable coarse structure from the first stage to correct fine-detail errors and generate high-quality try-on results.
}
  \label{fig:Pipeline_Overview}
\end{figure}
\section{Methods}
\label{sec:method}

\paragraph{Notations.}
Given a person image \( \mathbf{x}_p \in \mathbb{R}^{H \times W \times 3} \) and a garment image \( \mathbf{x}_g \in \mathbb{R}^{H \times W \times 3} \), the virtual try-on task generates a realistic output \( \mathbf{x}_r \in \mathbb{R}^{H \times W \times 3} \), where the person wears the given garment.  
In the low-resolution stage, the inputs \( \mathbf{x}_p \) and \( \mathbf{x}_g \) are downsampled to \( \tilde{\mathbf{x}}_p, \tilde{\mathbf{x}}_g \in \mathbb{R}^{h \times w \times 3} \), with \( h = H/\sigma \), \( w = W/\sigma \), and \( \sigma \) denoting the downsampling ratio. These are used to generate a low-resolution result \( \tilde{\mathbf{x}}_r \in \mathbb{R}^{h \times w \times 3} \).  
For simplicity, we assume that \( \tilde{\mathbf{x}}_r \) is upsampled to the original resolution in the high-resolution stage, and do not introduce a separate symbol.

\paragraph{Mask-free strategy.}
Previous virtual try-on methods~\citep{kim2024stableviton,xu2025ootdiffusion,choi2025improving,zhou2025flowattention,jiang2024fitdit} typically use external human parsers to generate body segmentation masks for garment localization.
However, in diffusion-based models like Stable Diffusion~\citep{rombach2022high}, which already encode strong human-structure priors, this is unnecessary.
We therefore adopt a fully mask-free design: the model directly consumes the garment and person images during training and inference, without parser-based pseudo-masks or segmentation guidance.

\subsection{Overview}
\label{subsec:overview}
Our method is built upon Stable Diffusion~\citep{rombach2022high}. The backbone architecture follows the dual U-Net framework~\citep{zhang2023referencecontrolnet,hu2024animate,xu2024magicanimate}, which has also demonstrated strong performance in virtual try-on tasks~\citep{choi2025improving,xu2025ootdiffusion}. Further architectural details are provided in Subsection~\ref{subsec:architecture}. As shown in Figure~\ref{fig:Pipeline_Overview}, both the low-resolution and high-resolution stages share this network architecture. 

In the low-resolution stage, images inherently emphasize structural information because fine-grained textures are suppressed. This stage therefore serves two main purposes. First, it enables the model to capture the human body shape and garment category, ensuring accurate structural alignment between the person and the clothing. Second, it provides coarse but semantically reliable garment structure, such as the placement of stripes or patterns, offering guidance for fine-detail reconstruction in the high-resolution stage.

In the high-resolution stage, the key challenge is how to effectively utilize the low-resolution result \(\tilde{\mathbf{x}}_r\). Here, we introduce our blend-refine diffusion process, which defines the noisy latent input as
\begin{equation}
\mathbf{x}_T = \alpha \cdot \boldsymbol{\epsilon} + \beta \cdot \tilde{\mathbf{x}}_r,
\label{eq:high_reso_x_T}
\end{equation}
where \(\boldsymbol{\epsilon} \sim \mathcal{N}(\mathbf{0}, \mathbf{I})\), and \(\alpha, \beta\) are balancing coefficients. The model is trained to gradually denoise this latent, transitioning \(\mathbf{x}_T\) to \(\mathbf{x}_r\) by converting the noise component \(\alpha \cdot \boldsymbol{\epsilon}\) into the residual term \(\mathbf{x}_r - \beta \cdot \tilde{\mathbf{x}}_r\), such that the final result satisfies \(\mathbf{x}_0 = \mathbf{x}_r\).
We detail the full formulation, as well as the design choices for \(\alpha\) and \(\beta\), in Subsection~\ref{subsec:high_res}.

\subsection{Network architecture}
\label{subsec:architecture}

We adopt a dual U-Net architecture, where the reference U-Net encodes garment features and integrates them into the main denoising U-Net via self-attention layers, a structure proven effective for garment fidelity and visual quality~\citep{choi2025improving}.
Following the approach of~\citep{chong2025catvton}, we remove all cross-attention layers, relying solely on self-attention, which improves performance and efficiency, as shown by our ablation studies (see Appendix~\ref{ablation:remove_cross}). 
To enhance computational efficiency, we execute the reference U-Net once per sample, using it solely as a conditioning module, as in~\citep{li2024anyfit}. 

We initialize our U-Net weights using those from Stable Diffusion 1.5~\citep{rombach2022high}.  
While SDXL~\citep{podell2025sdxl} offers greater generative power, our goal is a lightweight yet effective framework, so we retain SD1.5 as the backbone.  
Preliminary experiments with transformer-based architectures like SD3 and SD3.5 are discussed in Appendix~\ref{sec:discussion}.  
In conclusion, SD1.5 strikes the optimal balance between model simplicity and try-on performance, forming the foundation of our pipeline.

\subsection{Low-resolution stage}
\label{subsec:low_res}

In this stage, we first downsample the garment image \(\mathbf{x}_g\) and the person image \(\mathbf{x}_p\) by the downsampling ratio \(\sigma\). 
The downsampled garment image \(\tilde{\mathbf{x}}_g\) is encoded by a VAE~\citep{kingma2013auto} and then fed into the reference U-Net. 
The downsampled person image \(\tilde{\mathbf{x}}_p\) is also VAE-encoded and concatenated along the feature channels with a Gaussian noise tensor of the same shape. 
This combined latent is passed to the denoising U-Net to generate a low-resolution result \(\tilde{\mathbf{x}}_r \in \mathbb{R}^{h \times w \times 3}\). 
Both training and inference follow the standard diffusion process used in Stable Diffusion~\citep{rombach2022high,ho2020denoising,song2020denoising}.

The only hyperparameter in this stage is the downsampling ratio \(\sigma\), with \(\sigma \in \{1,2,4\}\).
As illustrated in Figure~\ref{fig:Pipeline_Overview}, when \(\sigma = 1\), the low-resolution stage operates at the same resolution as the high-resolution stage, \textbf{violating the purpose of leveraging lower resolution for improved structural modeling and introducing artifacts due to difficulty}.
\(\sigma = 2\) corresponds to downsampling from \(768 \times 1024\) to \(384 \times 512\), and so on.
Our observations show that both \(\sigma = 2\) and \(\sigma = 4\) enhance human-body structural understanding, \textbf{but, as also evident in Figure~\ref{fig:Pipeline_Overview}, \(\sigma = 4\) sacrifices structural detail}, making \(\sigma = 2\) the most reliable choice for accurate garment reconstruction. Quantitative and qualitative comparisons across different \(\sigma\) values are provided in Subsection~\ref{ablation}, and \(\sigma = 2\) is adopted for all experiments in this work.

\subsection{High-resolution stage}
\label{subsec:high_res}

In this stage, \(\mathbf{x}_g\) and \(\mathbf{x}_p\) are used in the same way as in the low-resolution stage: \(\mathbf{x}_g\) is passed to the reference U-Net, while \(\mathbf{x}_p\) is encoded and concatenated with the latent, which is then input to the denoising U-Net. The main difference from the low-resolution stage lies in the initialization of the latent and the denoising process.

\subsubsection{Reformulating the denoising process with blend-refine}

\paragraph{DDPM.}
Denoising Diffusion Probabilistic Mode~(DDPM)~\citep{ho2020denoising} aims to approximate the true data distribution by leveraging the diffusion probabilistic model framework~\citep{sohl2015deep}, which defines a Markov chain of length \(T\) that gradually transforms pure Gaussian noise into a sample from the data distribution.
Compared with earlier diffusion models, DDPM incorporates ideas from score matching~\citep{song2019scorematching}, simplifying the objective by training the model to predict only the noise component \(\boldsymbol{\epsilon}\), which approximates the score function (\textit{i.e.}, the gradient of the log-density). 
The forward and reverse diffusion processes are defined as:
\begin{align}
\mathbf{x}_t &= \sqrt{\alpha_t} \, \mathbf{x}_{t-1} + \sqrt{1 - \alpha_t} \, \boldsymbol{\epsilon} \notag \\
             &= \sqrt{\bar{\alpha}_t} \, \mathbf{x}_0 + \sqrt{1 - \bar{\alpha}_t} \, \boldsymbol{\epsilon} \label{eq:DDPM_forward}, \\
\mathbf{x}_{t-1} &= \frac{1}{\sqrt{\alpha_t}} \left( \mathbf{x}_t - \frac{1 - \alpha_t}{\sqrt{1 - \bar{\alpha}_t}} \, \boldsymbol{\epsilon}_\theta(\mathbf{x}_t, t) \right) + \sigma_t \mathbf{z}. \label{eq:DDPM_reverse}
\end{align}
\Eqref{eq:DDPM_forward} and~\Eqref{eq:DDPM_reverse} define the forward and reverse diffusion processes, respectively. In these equations, \(\boldsymbol{\epsilon}, \mathbf{z} \sim \mathcal{N}(\mathbf{0}, \mathbf{I})\), and \(\boldsymbol{\epsilon}_\theta(\mathbf{x}_t, t)\) denotes the noise predicted by the model. The parameters \(\alpha_t\) and \(\bar{\alpha}_t\) are predefined noise scheduling terms. In this framework, \(\mathbf{x}_0 \sim p_{data}(\mathbf{x}_0)\) denotes a sample from the true data distribution, while \(\mathbf{x}_T \sim \mathcal{N}(\mathbf{0}, \mathbf{I})\). All formulations above hold for \(t = 1, 2, \dots, T\).

\paragraph{Blend-refine diffusion reformulation.}
Inspired by~\citep{yue2023resshift}, we reformulate the forward and reverse processes to \textbf{explicitly} incorporate the low-resolution result \(\tilde{\mathbf{x}}_r\) as a structural prior for high-resolution generation.  
\textbf{Rather than constructing a Markov chain only between Gaussian noise and \(p_{\mathrm{data}}(\mathbf{x}_0)\), we aim to build a transition path from \(\tilde{\mathbf{x}}_r\) to the high-resolution result \(\mathbf{x}_r\).}  
In the original formulation of~\citep{yue2023resshift}, \(\mathbf{x}_T = \boldsymbol{\epsilon} + \tilde{\mathbf{x}}_r\) with \(\mathbf{x}_0 = \mathbf{x}_r\), which offers limited flexibility to control the relative influence of \(\tilde{\mathbf{x}}_r\) and \(\boldsymbol{\epsilon}\).  
To overcome this limitation, we introduce a simple extension with two coefficients for more flexible initialization, as defined in \Eqref{eq:high_reso_x_T}, while keeping \(\mathbf{x}_0 = \mathbf{x}_r\).  
\textbf{This straightforward modification yields significant improvements, as shown in Subsection~\ref{ablation}.}  
Under this formulation, the blend-refine forward and reverse diffusion processes become:
\begin{align}
\mathbf{x}_t &= \sqrt{\alpha_t} \, \mathbf{x}_{t-1} + \sqrt{1 - \alpha_t} \left( \alpha \cdot \boldsymbol{\epsilon} + \beta \cdot \tilde{\mathbf{x}}_r \right) \notag \\
             &= \sqrt{\bar{\alpha}_t} \, \mathbf{x}_0 + \sqrt{1 - \bar{\alpha}_t} \left( \alpha \cdot \boldsymbol{\epsilon} + \beta \cdot \tilde{\mathbf{x}}_r \right), \label{eq:residual_forward} \\
\mathbf{x}_{t-1} &= \frac{1}{\sqrt{\alpha_t}} \left( \mathbf{x}_t - \frac{1 - \alpha_t}{\sqrt{1 - \bar{\alpha}_t}} \, \tilde{\boldsymbol{\epsilon}}_\theta(\mathbf{x}_t, t) \right) + \sigma_t \mathbf{z}. \label{eq:residual_reverse}
\end{align}
\Eqref{eq:residual_forward} and~\Eqref{eq:residual_reverse} define the forward and reverse processes, respectively. The model is trained to predict \(\tilde{\boldsymbol{\epsilon}}_\theta(\mathbf{x}_t, t) \approx \alpha \cdot \boldsymbol{\epsilon} + \beta \cdot \tilde{\mathbf{x}}_r\), such that the noise component in \(\mathbf{x}_t\) is replaced by a composition of \(\boldsymbol{\epsilon}\) and the low-resolution result \(\tilde{\mathbf{x}}_r\). 
Except for this reformulated initialization and noise target, the rest of the denoising process remains consistent with the original DDPM. All formulations above hold for \(t = 1, 2, \dots, T\).

\subsubsection{Controlling noise-structure balance}
In the high-resolution stage, generation starts from the latent
\(\mathbf{x}_T = \alpha \cdot \boldsymbol{\epsilon} + \beta \cdot \tilde{\mathbf{x}}_r\),
a combination of stochastic noise and a structural prior.
Here, \(\alpha\) controls randomness and \(\beta\) controls structural guidance.
\textbf{Intuitively, \(\beta\) should reflect the similarity between the two-scale distributions,
while \(\alpha\) should correspond to the size of the probability space
associated with the residual term \(\mathbf{x}_r - \beta \cdot \tilde{\mathbf{x}}_r\).}
As illustrated in Figure~\ref{fig:Pipeline_Overview}, an excessively large \(\beta\) forces overreliance on the low-resolution input,
suppressing fine-detail recovery, whereas a large \(\alpha\) introduces noise that can disrupt structure.
Empirically, setting \(\alpha = \beta = 0.5\) provides a balanced trade-off between fidelity and flexibility.
Further discussion is provided in Subsection~\ref{ablation}.

\section{Experiments}

\subsection{Experimental setup}
\label{subsec:experiment}

\paragraph{Datasets.}
In this paper, we conduct experiments on the VITON-HD~\citep{choi2021viton} and DressCode~\citep{morelli2022dress} datasets. All ablation studies are carried out on VITON-HD. While VITON-HD contains only upper-body garments, DressCode includes three garment categories: upper-body, lower-body, and dresses. Both datasets consist of paired images, each containing a person image and a corresponding garment image.
However, as our method is mask-free, the original paired data alone is insufficient for training. To address this, we use IDM-VTON~\citep{choi2025improving} to synthesize additional training data. Further details are provided in Appendix~\ref{subsec:dataset}.

\paragraph{Implementation details.}
For network initialization, both the reference U-Net and the denoising U-Net are initialized with pretrained weights from Stable Diffusion 1.5~\citep{rombach2022high}. As detailed in Subsections~\ref{subsec:low_res} and~\ref{subsec:high_res}, we set the downsampling ratio \(\sigma = 2\) and use \(\alpha = \beta = 0.5\) to initialize \(\mathbf{x}_T\) in the high-resolution stage. Consequently, the low-resolution stage operates at a resolution of \(384 \times 512\), while the high-resolution stage produces outputs at \(768 \times 1024\).
During inference, the two stages are executed sequentially with 20 sampling steps each, using the DDIM sampler~\citep{song2020denoising}. Additional training details are provided in Appendix~\ref{subsec:more_implementation}.

\paragraph{Baselines.}
We compare our method with several recent state-of-the-art approaches, including CatVTON~\citep{chong2025catvton}, IDM-VTON~\citep{choi2025improving}, Leffa~\citep{zhou2025flowattention}, OOTDiffusion~\citep{xu2025ootdiffusion}, and FitDiT~\citep{jiang2024fitdit}, using their official model weights and inference code. For fairness, we standardize the number of inference steps to 30 across all methods.
All methods, except FitDiT, are trained separately on each dataset and evaluated accordingly. FitDiT provides only a single set of pretrained weights, jointly trained on VITON-HD, DressCode, and CVDD~\citep{jiang2024fitdit}, and is included in our evaluation for completeness.

\paragraph{Evaluation metrics.}
Previous virtual try-on methods typically evaluate performance under both paired and unpaired settings. The paired setting involves reconstructing the original person image with the same garment, while the unpaired setting replaces it with a different one~\citep{choi2021viton}. As most prior approaches rely on masking the garment region, they support both settings. In contrast, our method is mask-free and is therefore evaluated only under the unpaired setting, which better reflects real-world scenarios.
We adopt Fréchet Inception Distance (FID)~\citep{parmar2022aliased} and Kernel Inception Distance (KID)~\citep{binkowski2018demystifying} as quantitative metrics. We also conduct a user study to assess perceptual quality: for VITON-HD, we randomly sample 100 results per method; for DressCode, we sample 33, 33, and 34 results from the dresses, lower-body, and upper-body categories, respectively. Participants are asked to select the result they think performs better in the try-on task. All evaluations are conducted at a resolution of \(768 \times 1024\).

\begin{figure}[t]
  \centering
  \includegraphics[width=\linewidth]{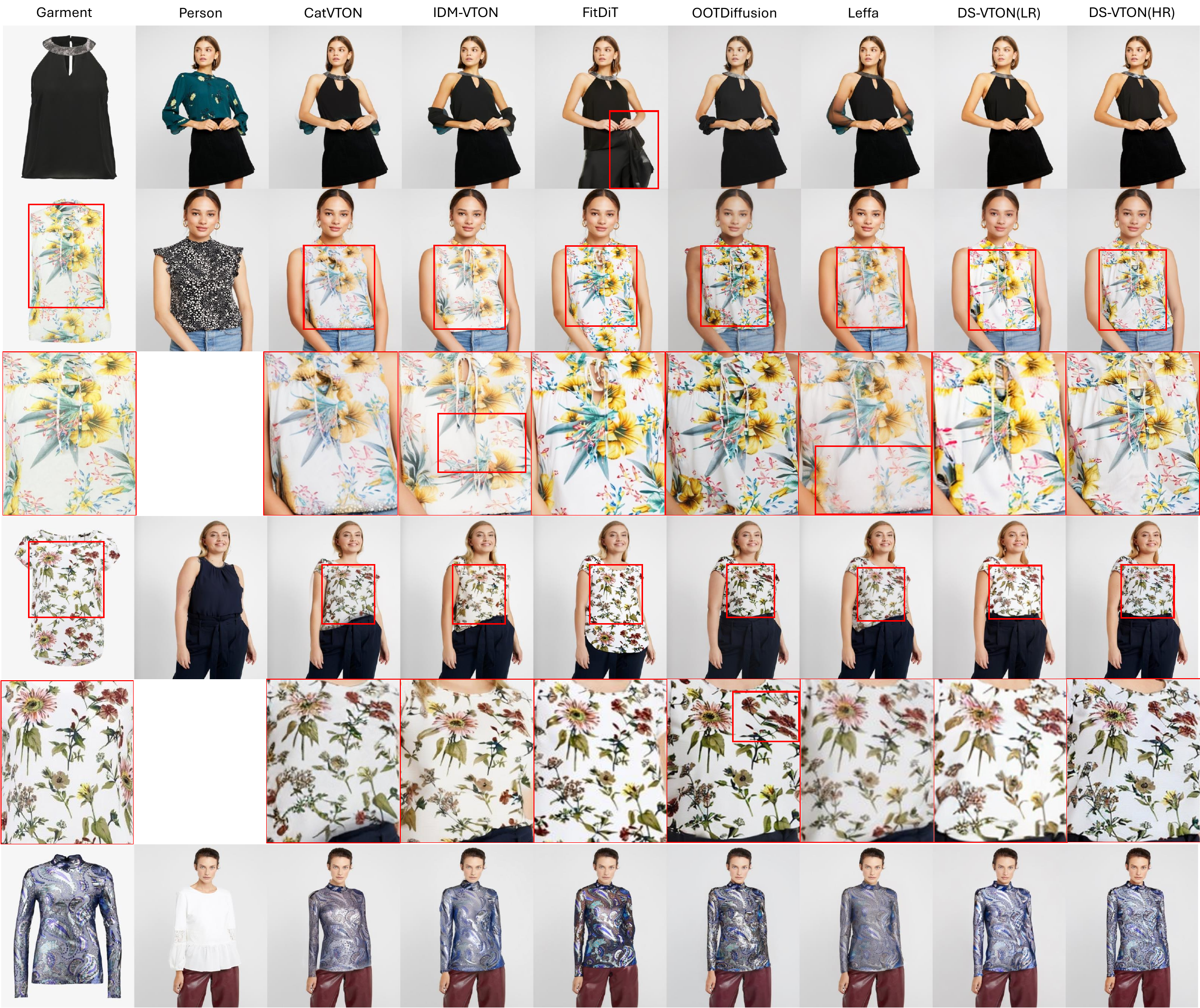}
  \caption{Qualitative comparison on the VITON-HD dataset. DS-VTON(LR) denotes the low-resolution result, and DS-VTON(HR) represents the final high-resolution result.}
  \label{fig:Compare}
\end{figure}

\subsection{Quantitative and qualitative results}
\label{subsec:compare}

\paragraph{Qualitative comparison.}
Figure~\ref{fig:Compare} presents a qualitative comparison between DS-VTON and recent baseline methods on the VITON-HD~\citep{choi2021viton} dataset. Row numbers referenced below correspond to Figure~\ref{fig:Compare}. We evaluate each method in terms of two key aspects: structural alignment and detail preservation.
In terms of structural alignment, most mask-based methods fail to accurately capture body pose and garment shape, as illustrated in Row~1. FitDiT~\citep{jiang2024fitdit}, which uses a larger rectangular mask, performs relatively better but introduces noticeable artifacts: it fails to reconstruct the hands and exhibits visible artifacts at the junction of the upper and lower garments in Row~1. OOTDiffusion~\citep{xu2025ootdiffusion} alters the original skin tone in Row~2. CatVTON~\citep{chong2025catvton}, IDM-VTON~\citep{choi2025improving}, and Leffa~\citep{zhou2025flowattention} also show varying degrees of misalignment. In contrast, DS-VTON consistently achieves accurate alignment across a wide range of poses and garment types.
Regarding detail preservation, CatVTON and IDM-VTON fail to retain fine-grained textures on complex garments (Row~3), with IDM-VTON generating oversimplified or missing patterns. While Leffa, OOTDiffusion, and FitDiT better preserve textures, they each show limitations: FitDiT achieves texture preservation at the cost of distorting the person’s actual body shape and dressing structure (Rows~2 and 4), OOTDiffusion introduces artifact patterns (Row~5), and Leffa exhibits reduced pattern clarity in complex regions (Row~3). Furthermore, both FitDiT and Leffa suffer from tonal inconsistencies—FitDiT produces noticeably brighter garments (Row~6), while Leffa tends to generate darker outputs. In contrast, DS-VTON preserves fine textures while maintaining tonal fidelity throughout.

\begin{table*}[t]
\caption{Quantitative comparisons on the VITON-HD~\citep{choi2021viton} and DressCode~\citep{morelli2022dress} datasets. FitDiT~\citep{jiang2024fitdit}, trained jointly on VITON-HD, DressCode, and CVDD~\citep{jiang2024fitdit}, is included for completeness. In contrast, all other methods are trained individually on each dataset.}
\centering
\fontsize{8pt}{9pt}\selectfont
\begin{tabular}{ccccccc}
  \toprule
  \multicolumn{1}{c}{\textbf{Dataset}} & \multicolumn{3}{c}{VITON-HD} & \multicolumn{3}{c}{DressCode} \\
  \cmidrule(lr){1-1}\cmidrule(lr){2-4} \cmidrule(lr){5-7}
  \multicolumn{1}{c}{\textbf{Method}}
  & FID ↓ & KID ↓ & User Study ↑ & FID ↓ & KID ↓ & User Study ↑\\
  \midrule
  OOTDiffusion~\citep{xu2025ootdiffusion} & \underline{9.02} & \underline{0.63} & 4.1 & 7.10 & 2.28 & 7.2 \\
  IDM-VTON~\citep{choi2025improving}            & 9.10  & 1.06 & 11.6 & 5.51 & 1.42 & 9.1 \\
  CatVTON~\citep{chong2025catvton}             & 9.40 & 1.27 & 3.4 & 5.24  &  1.21 & 5.2 \\
  Leffa~\citep{zhou2025flowattention}            & 9.38  & 0.92 & 4.7 & 6.17 & 1.90 & 7.5 \\
  FitDiT~\citep{jiang2024fitdit}            & 9.33  & 0.89 & \underline{19.7} & \underline{4.47} & \underline{0.41} & \underline{34.3} \\
  \midrule
  \textbf{DS-VTON (ours)}                   & \textbf{8.24} & \textbf{0.31} & \textbf{56.5} & \textbf{4.21} & \textbf{0.34} & \textbf{36.7} \\
  \bottomrule
\end{tabular}
\label{tab:viton}
\end{table*}

\begin{table*}[t]
\centering
\fontsize{8pt}{9pt}\selectfont
\renewcommand{\arraystretch}{1.3}
\setlength{\tabcolsep}{3mm}

\begin{minipage}{0.48\linewidth}
  \centering
  \caption{Ablation study on dual-scale design and downsampling ratio \(\sigma\).}
  \label{tab:ablation_downsampling_ratio}
  \begin{tabular}{ccc}
    \toprule
    Version & FID~\(\downarrow\) & KID~\(\downarrow\) \\
    \midrule
    \(\sigma = 1\) & 8.97 & 1.01 \\
    \(\sigma = 1, \alpha=\beta=\tfrac{1}{2}\) & 8.77 & 0.61 \\
    \(\sigma = 4, \alpha=\beta=\tfrac{1}{2}\) & 8.41 & 0.57 \\
    \boldmath{\(\sigma = 2, \alpha=\beta=\tfrac{1}{2}\)} & \textbf{8.24} & \textbf{0.31} \\
    \bottomrule
  \end{tabular}
\end{minipage}
\hfill
\begin{minipage}{0.48\linewidth}
  \centering
  \caption{Ablation study on coefficients \(\alpha, \beta\) under fixed \(\sigma=2\).}
  \label{tab:alpha_beta}
  \begin{tabular}{ccc}
    \toprule
    Version & FID~\(\downarrow\) & KID~\(\downarrow\) \\
    \midrule
    \boldmath{\(\sigma = 2, \alpha=\beta=\tfrac{1}{2}\)} & \textbf{8.24} & \textbf{0.31} \\
    \(\sigma = 2, \alpha=\tfrac{2}{3}, \beta=\tfrac{1}{3}\) & 8.46 & 0.55 \\
    \(\sigma = 2, \alpha=\tfrac{1}{3}, \beta=\tfrac{2}{3}\) & 8.26 & 0.35 \\
    \(\sigma = 2, \alpha=\beta=1\) & 8.75 & 0.94 \\
    \bottomrule
  \end{tabular}
\end{minipage}
\end{table*}

\paragraph{Quantitative comparison.}
We conduct experiments on both the VITON-HD~\citep{choi2021viton} and DressCode~\citep{morelli2022dress} datasets. As shown in Table~\ref{tab:viton}, DS-VTON achieves substantial improvements across both benchmarks.
CatVTON~\citep{chong2025catvton} generates images at \(384 \times 512\), which we upsample to \(768 \times 1024\) for comparison. To confirm that this does not bias results, we also evaluate it at native resolution, obtaining FID and KID scores of 9.36 and 1.19, respectively—indicating minimal degradation due to upsampling.
Unlike prior methods~\citep{chong2025catvton,zhou2025flowattention,choi2025improving} that rely on explicit mask generation and inpainting, DS-VTON is entirely mask-free. This enables accurate, high-quality outputs that are robust and reproducible without dependence on mask quality.

\subsection{Ablation study}
\label{ablation}

\begin{figure}[t]
  \centering
  \includegraphics[width=\linewidth]{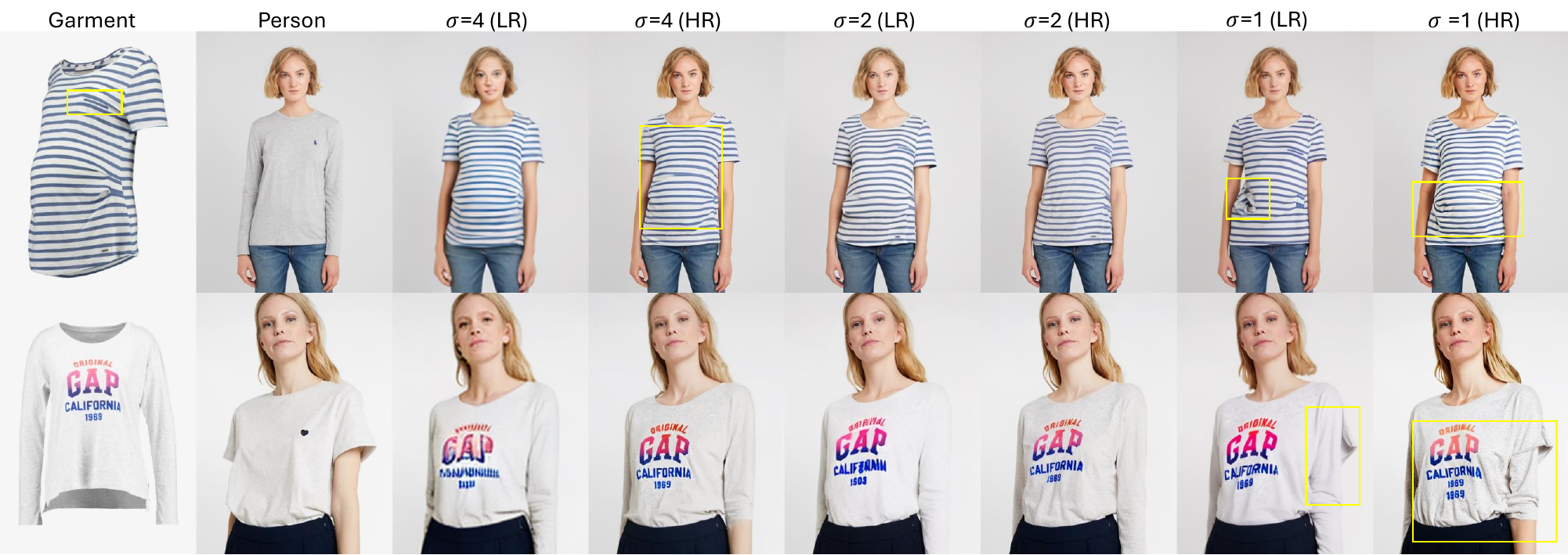}
  \caption{Visualized results under varying downsampling ratios $\sigma$.}
  \label{fig:downsampling_ratio}
\end{figure}

\begin{figure}[t]
  \centering
  \includegraphics[width=\linewidth]{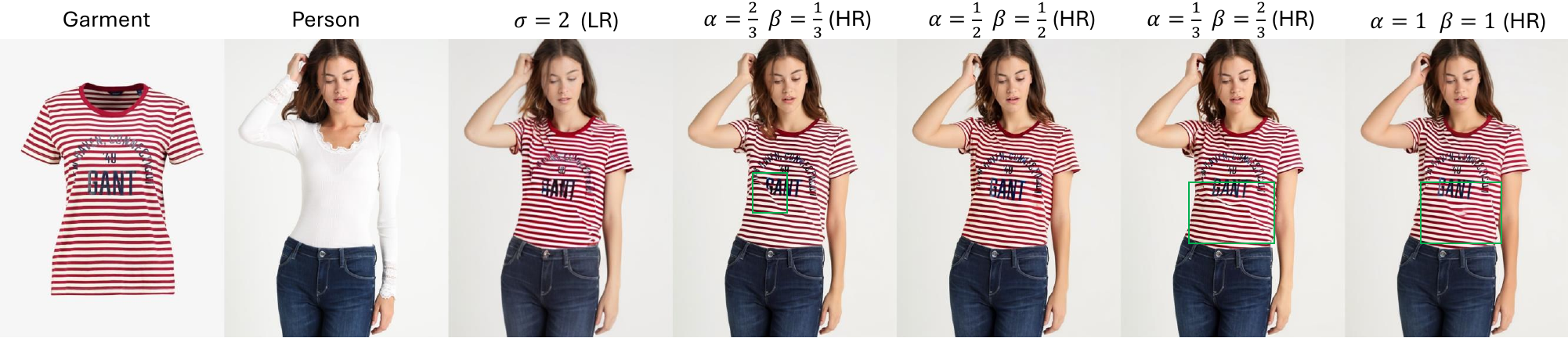}
  \caption{Visualized results under different \(\mathbf{x}_T\) initialization settings (\(\mathbf{x}_T = \alpha \cdot \boldsymbol{\epsilon} + \beta \cdot \tilde{\mathbf{x}}_r\)).}
  \label{fig:alpha_beta}
\end{figure}

Here we present several ablation studies to validate the rationale of our design. \textbf{In addition, we evaluate an alternative refinement method used in SDXL; see Subsection~\ref{sec:discussion} for details.}

\paragraph{Ablation on dual-scale design.}
To validate the effectiveness of our dual-scale design, we train four variants on the VITON-HD dataset. As shown in Table~\ref{tab:ablation_downsampling_ratio}, Row~1 (\(\sigma=1\)) corresponds to training the model directly at high resolution (\(768 \times 1024\)) without a refinement stage, aligning with the \(\sigma=1\) (LR) column in Figure~\ref{fig:downsampling_ratio}. \textbf{Applying the mask-free strategy at this resolution leads to poor structural results, highlighting the need for coarse-level guidance.}
Row~2 (\(\sigma=1, \alpha=\beta=\frac{1}{2}\)), shown in the \(\sigma=1\) (HR) column, adds a refinement stage. While some structural errors are alleviated, relying solely on the second stage to recover both structure and detail still results in failures, due to the lack of reliable low-resolution guidance. This also violates our design principle of decoupling structure modeling (stage one) from detail refinement (stage two).
Rows~3 and 4 evaluate the full dual-scale pipeline with \(\sigma = 4\) and \(\sigma = 2\), respectively. 
When garments are structurally simple, both settings perform reasonably well. 
However, as shown in Row~1 of Figure~\ref{fig:downsampling_ratio}, \textbf{\(\sigma = 4\) introduces visible information loss in complex cases. }
These qualitative results align with the quantitative trends: among the four variants, the single stage (\(\sigma = 1\)) and the dual stage with \(\sigma = 1\), \(\sigma = 2\), or \(\sigma = 4\), the single stage model performs worst, \(\sigma = 4\) shows moderate improvement, and \(\sigma = 2\) achieves the best overall performance.

\paragraph{Ablation on different initializations of \(\mathbf{x}_T\).}
When fixing the downsampling ratio to \(\sigma = 2\), the coefficients \(\alpha\) and \(\beta\) determine how strongly the low-resolution output influences the high-resolution stage.
As shown in Figure~\ref{fig:alpha_beta}, setting \(\alpha\) too high, e.g., \(\alpha=1,\beta=1\) or \(\alpha=\tfrac{2}{3},\beta=\tfrac{1}{3}\), causes structural distortions in the final result even though the low-resolution output already provides accurate guidance. In both cases the red stripe on the garment is distorted, and in the \(\alpha=\tfrac{2}{3},\beta=\tfrac{1}{3}\) configuration the ``GANT'' text also becomes warped.
\textbf{We attribute this to excessive randomness in the high-resolution stage, which then attempts to re-establish structure that should have been fully resolved in the low-resolution stage.}  
Among the remaining settings, \(\alpha=\tfrac{1}{3},\beta=\tfrac{2}{3}\) and \(\alpha=\tfrac{1}{2},\beta=\tfrac{1}{2}\) produce generally good results.  
However, the \(\alpha=\tfrac{1}{3},\beta=\tfrac{2}{3}\) setting still exhibits minor distortion in the red stripe, indicating insufficient restoration of fine details. \textbf{Here, the low-resolution result exerts too strong an influence: while structurally accurate, it lacks fine details, and the high \(\beta\) prevents the high-resolution stage from effectively correcting those errors.}
Based on these observations, we adopt \(\alpha = \beta = \tfrac{1}{2}\) as our default setting.
The quantitative results in Table~\ref{tab:alpha_beta} support this choice.

\section{Conclusions}
We propose DS-VTON, a dual-scale framework that employs blend-refine denoising to bridge low- and high-resolution generation. 
This design enables a more effective coarse-to-fine process and, combined with a mask-free strategy, achieves significant improvements in both visual quality and robustness over existing methods. 
Furthermore, the paradigm is inherently scalable and generalizable, with clear potential for extension to higher resolutions and to broader generation tasks beyond virtual try-on, such as personalized image synthesis.

\bibliographystyle{plainnat}
\bibliography{neurips_2025}

\appendix
\section{Appendix}

\subsection{Experimental details}
\subsubsection{Additional training data generation and associated challenges}
\label{subsec:dataset}
Our method adopts a mask-free paradigm, in contrast to prior approaches~\citep{choi2025improving,xu2025ootdiffusion,li2024anyfit}, which rely on paired person-garment images by masking out the garment region in the person image and reconstructing it using the standalone garment image. In our case, the person image remains unaltered throughout the training process.
To train the low-resolution stage, each training sample requires three images: (1) a garment image, (2) a person image wearing that garment, and (3) another person image of the same identity but wearing a different garment. The third image is constructed by randomly sampling another garment of the same category from the dataset and synthesizing a new person image using IDM-VTON~\citep{choi2025improving}.
For the high-resolution stage, we additionally require the low-resolution output corresponding to the original person-garment pair. To obtain this, we use our trained low-resolution model to generate a coarse try-on result by inputting the target garment and the synthesized person image (i.e., with a different garment). This output is then used as the low-resolution input for the high-resolution stage.

In summary, each high-resolution training sample consists of: (1) the target garment image, (2) a person image wearing a different garment, (3) the corresponding low-resolution try-on result for the target garment, and (4) the ground-truth high-resolution image of the person wearing the target garment.

While the above construction enables mask-free supervision, it inevitably introduces certain artifacts. Specifically, in the low-resolution stage, the reference U-Net encodes the garment image, while the denoising U-Net operates on the concatenation of the noisy latent and the person image—here, a synthesized image of the same identity wearing a different garment.

Since the synthesized person image is generated by a model such as IDM-VTON, it may exhibit variations beyond the garment itself, including changes in hairstyle, background, or the presence of accessories. As a result, the model may inadvertently learn to alter these unrelated regions during training. Ideally, such issues would not occur if fully disentangled and clean ground-truth data were available.
Fortunately, we find this side effect to be limited in practice, as most synthesized person images remain visually consistent with the original identity.

\subsubsection{More implementation details}
\label{subsec:more_implementation}
All experiments are conducted on 8 NVIDIA A6000 GPUs. For both VITON-HD~\citep{choi2021viton} and DressCode~\citep{morelli2022dress}, the low-resolution stage is trained with a batch size of 8, and the high-resolution stage with a batch size of 2. Both stages are optimized using the AdamW optimizer~\citep{loshchilov2017fixing} with a learning rate of \(1\text{e}{-6}\).
For VITON-HD, the low-resolution and high-resolution stages are trained for 15,000 and 30,000 steps, respectively (approximately 5 and 24 hours). For DressCode, we jointly train all three garment categories, with the two stages trained for 20,000 and 30,000 steps (approximately 7 and 24 hours).
During inference, both stages use 20 DDIM sampling steps.  
On a single A6000 GPU, the low-resolution stage takes about 1 second per sample, while the high-resolution stage takes about 4 seconds.  
Although our method includes two stages, the low-resolution stage is much faster than the high-resolution stage, so the total runtime remains acceptable.  
\textbf{Compared with the methods in our earlier comparison, our approach is only slower than CatVTON, comparable to FitDiT, and faster than the others.}

\subsubsection{Additional Comparisons with Other Methods}
We detail here the data requirements of the methods introduced in Subsection~\ref{subsec:experiment} to clarify the comparisons.  
For completeness, we also present paired-setting results on VITON-HD.  
All competing methods except ours require an agnostic mask; Leffa additionally uses a DensePose map; FitDiT requires human body keypoints; and IDM-VTON depends on both detailed garment descriptions and a DensePose map.  
Our method, by contrast, is trained only on paired images of the same person wearing different garments.

Because our approach does not mask the original garment region, the input and expected output under the paired setting are identical.  
Nevertheless, we include two evaluation variants:
\begin{itemize}[leftmargin=*]
\item \textbf{DS-VTON (direct)}: directly uses the garment and person images to generate the result.
\item \textbf{DS-VTON (train-way)}: first employs IDM-VTON to create an image of the person wearing another garment, then applies our method to re-dress the original garment.  
This procedure involves two rounds of unpaired try-on and inevitably introduces additional artifacts, so it does not fully reflect our method’s capability.
\end{itemize}
The quantitative results appear in Table~\ref{tab:paired}.

\subsection{More ablation studies}
\subsubsection{Ablation on removing cross-attention layers in U-Net}
\label{ablation:remove_cross}

\begin{figure}[t]
  \centering
  \includegraphics[width=\linewidth]{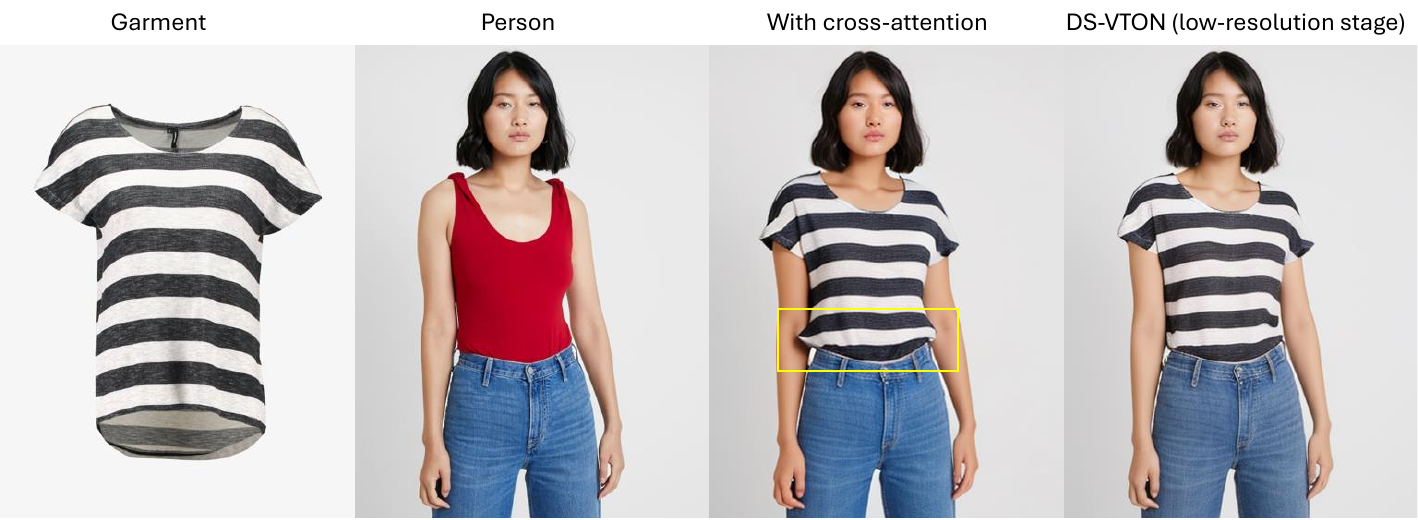}
  \caption{Comparison between our baseline architecture and the variant with cross-attention. In the variant, garment features encoded by CLIP are injected into both the reference U-Net and the denoising U-Net via cross-attention layers.}
  \label{fig:remove_cross}
\end{figure}

Several previous try-on methods~\citep{choi2025improving,xu2025ootdiffusion,li2024anyfit} based on dual U-Net architectures incorporate additional conditional encoders, such as CLIP~\citep{radford2021learning} or IP-Adapter~\citep{ye2023ipadapter}, to inject garment information via cross-attention. However, these approaches introduce extra complexity, and it remains unclear whether they actually improve performance.
To investigate this, we conduct an ablation study on the VITON-HD~\citep{choi2021viton} dataset by comparing two versions of our low-resolution stage (\(384 \times 512\)): (1) our baseline model, which does not include any cross-attention layers, and (2) a variant built on the baseline by adding cross-attention layers to both the reference U-Net and the denoising U-Net. In the latter, garment features are encoded using CLIP and injected via cross-attention.
As shown in Figure~\ref{fig:remove_cross} and Table~\ref{tab:ablation_remove_cross}, adding cross-attention does not help preserve garment details; on the contrary, it introduces noticeable distortions. This is also reflected in the performance metrics.

\begin{table}[t]
\centering
\caption{Quantitative comparisons on the VITON-HD dataset under the paired setting.}
\label{tab:paired}
\begin{tabular}{lcccc}
\toprule
Method & FID~\(\downarrow\) & KID~\(\downarrow\) & SSIM~\(\uparrow\) & LPIPS~\(\downarrow\) \\
\midrule
OOTDiffusion       & 6.47 & 1.24 & 0.88 & 0.08 \\
IDM-VTON           & 5.84 & 0.77 & 0.87 & 0.06 \\
CatVTON            & 5.70 & 0.50 & 0.88 & 0.09 \\
Leffa             & 5.76 & 0.55 & 0.89 & 0.06 \\
FitDiT            & 7.27 & 0.73 & 0.84 & 0.09 \\
\textbf{DS-VTON (direct)}   & \textbf{4.75} & 0.43 & \textbf{0.90} & \textbf{0.05} \\
DS-VTON (train-way) & 5.23 & \textbf{0.31} & 0.89 & 0.06 \\
\bottomrule
\end{tabular}
\end{table}

\begin{table*}[t]
  \centering
  \caption{Ablation study on the effect of cross-attention layers in U-Net.}
  \label{tab:ablation_remove_cross}
  \begin{tabular}{ccc}
    \toprule
    Version & FID~\(\downarrow\) & KID~\(\downarrow\) \\
    \midrule
    With cross-attention & 9.07 & 0.94 \\
    \textbf{DS-VTON (low-resolution stage)} & \textbf{8.88} & \textbf{0.72} \\
    \bottomrule
  \end{tabular}
\end{table*}

\subsection{Discussions}
\label{sec:discussion}

\begin{figure}[t]
  \centering
  \includegraphics[width=\linewidth]{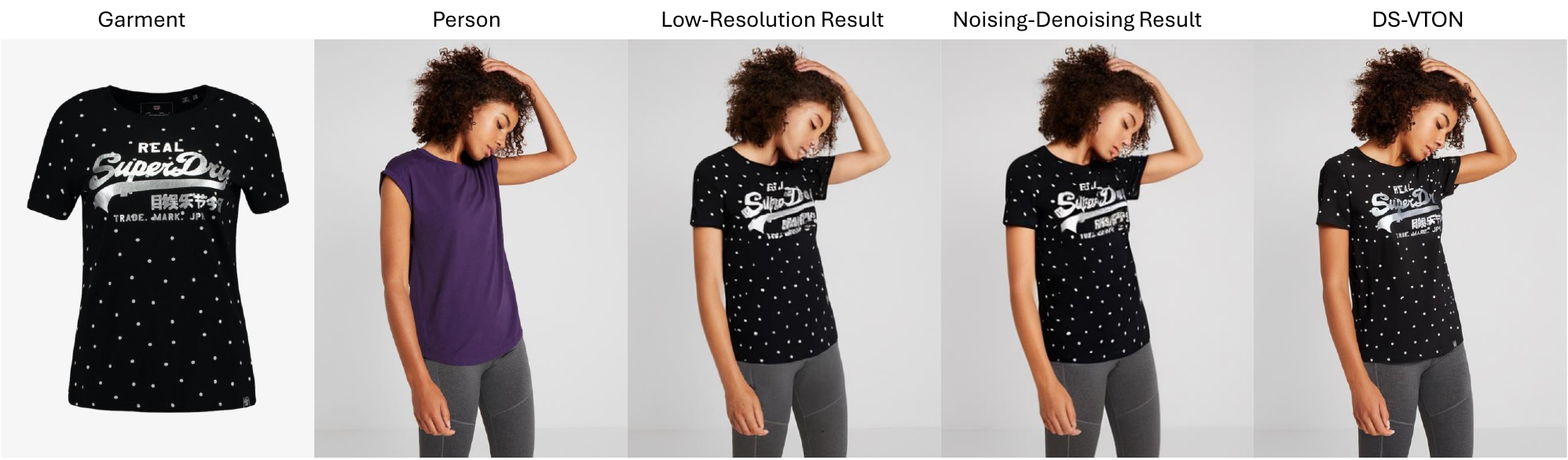}
  \caption{Comparison of qualitative results between our baseline and the SDXL variant. Quantitatively, the SDXL variant achieves an FID/KID of \textbf{8.98/0.65}, while our DS-VTON achieves \textbf{8.24/0.31}.}
  \label{fig:refinement}
\end{figure}

\paragraph{How about utilizing the DiT architecture for mask-free try-on?}
We also explored using the DiT~\citep{peebles2023scalable} architecture for mask-free try-on. Specifically, we implemented two variants based on SD3 and SD3.5~\citep{esser2024rectifiedflow}, constructing a dual-DiT structure analogous to the dual U-Net design. The reference DiT encodes the garment image, while the person image is concatenated with the latent tensor along the sequence dimension and passed to the denoising DiT.
To integrate the two branches, we follow a structure-aligned fusion strategy: since the reference DiT and denoising DiT share the same architecture, we concatenate the latent features from the corresponding transformer block of the reference DiT into the denoising DiT block, specifically on the key and value (K, V) inputs of the attention layer, before computing self-attention. We also remove the text encoder input entirely, so the joint attention layers in DiT degenerate into pure self-attention.
However, both versions failed to converge during training. We speculate that directly applying DiT to the mask-free try-on task may be suboptimal, and therefore did not pursue further investigation.

\paragraph{How about adopting a refinement mechanism similar to that of SDXL?}
We refer to the refinement method in SDXL~\citep{podell2025sdxl} as the SDXL strategy. During training, a separate model is trained to denoise only the final 200 steps of the high-resolution diffusion process. At inference, a low-resolution result is generated, upsampled, and injected with Gaussian noise at timestep 200, after which the refinement model denoises to produce the final output.  
Although this method can yield visually plausible results for simple patterns, its overall performance is inferior, as shown in Figure~\ref{fig:refinement}.
\textbf{The reason, we believe, lies in the difference between data distributions involved.} Although our blend-refine diffusion process also begins with adding noise, it differs fundamentally in how it relates the two stages. In SDXL, the refinement model learns to denoise samples drawn solely from the high-resolution distribution, and it lacks an explicit mechanism to relate this with the low-resolution result. In contrast, our blend-refine diffusion bridges the gap between the low-resolution and high-resolution distributions. This connection is key: it allows the model to better capture the transformation between the two distributions involved. 
Importantly, while the original diffusion process constructs a mapping from a simple, tractable distribution (such as Gaussian noise) to a complex data distribution, the blend-refine diffusion builds a direct bridge between two complex distributions. 

\paragraph{Difference between Our Blend-Refine Process and the DCI-VTON Refinement Branch.}
DCI-VTON also adds noise to the first-stage result and attempts to recover the final output in its second-stage refinement branch.
However, it does not explicitly transition the data distribution from the first stage to the second.  
Its second stage contains two branches: a refinement branch and a reconstruction branch.  
In the refinement branch, when the added noise is weak, the noisy input is nearly identical to the first-stage result, so the denoised output remains overly similar to the input and struggles to reach the desired final distribution.  
This is why \textbf{DCI-VTON requires two branches} and adopts a perceptual (VGG) loss rather than a pixel-wise loss in the refinement branch.  
The refinement branch provides only implicit guidance, while the reconstruction branch handles explicit Gaussian noise prediction.
By contrast, DS-VTON needs only a single branch.  
Our blend-refine denoising explicitly transfers the distribution from coarse to fine by altering the noise formulation itself, rather than relying on pure Gaussian noise.  
This change enables a principled transition between the low- and high-resolution distributions and provides stronger controllability—an essential design choice of our dual-scale framework.
To illustrate, suppose we follow DCI-VTON’s approach.  
Let \(\tilde{\mathbf{x}}_r\) denote the low-resolution output, \(\boldsymbol{\epsilon}\) the noise, and \(\mathbf{x}_r\) the high-resolution ground truth. From a data distribution perspective, if we adopt pure Gaussian noise as the only perturbation mechanism, there is a fundamental mismatch between the training and inference processes. During training, at any timestep $t$, the noisy latent is constructed by adding noise to $\tilde{\mathbf{x}}_r$, i.e., it is solely a function of $\tilde{\mathbf{x}}_r$ and $\boldsymbol{\epsilon}$. However, this setup does not align with inference. In inference, regardless of how the latent is initialized (whether from pure Gaussian noise or some other distribution), the moment a single denoising step is performed, the latent becomes correlated with the target distribution $\mathbf{x}_r$. This stands in stark contrast to training, where the noisy latent at each timestep is always constructed based on $\tilde{\mathbf{x}}_r$.
\textbf{The key difference is that we modify the noise formulation:  
rather than using pure Gaussian noise, our blend-refine denoising bridges the low- and high-resolution data distributions,  
providing strong controllability and serving as a fundamental design choice in our dual-scale framework.}

\paragraph{Broader impacts.}
The ability of DS-VTON to generate realistic virtual try-on results at multiple resolutions makes it well-suited for practical deployment in e-commerce scenarios, where different resolutions are often required across platforms and devices. At the same time, as with other generative technologies, DS-VTON may raise concerns related to intellectual property and personal privacy. We encourage its responsible and ethical use.

\paragraph{Limitation and Future Work.}
As discussed in Subsection~\ref{subsec:dataset}, one key limitation of our method lies in the data generation process. Due to the reliance on synthesized person images, the model may inadvertently learn to alter regions unrelated to the garment (e.g., hair, accessories, or background). While this issue is not severe in most cases, we acknowledge it as a limitation and consider improving data disentanglement and identity preservation an important direction for future work.
Another limitation stems from the use of fixed coefficients \(\alpha\) and \(\beta\) to initialize the high-resolution refinement stage. Although this static strategy proves effective, it may be overly rigid. In future work, we plan to investigate adaptive or learnable coefficient scheduling mechanisms, which could offer more flexible and content-aware refinement during generation.

\subsection{The Use of Large Language Models}
The Large Language Model was used solely for refining the text and improving the clarity and readability of the paper. It did not contribute to the research ideation or experimental design.

\subsection{More experiment results}
In this section, we present additional qualitative comparisons on the DressCode~\citep{morelli2022dress} dataset, more results on the VITON-HD~\citep{choi2021viton} dataset, and additional in-the-wild examples.

\begin{figure}[h]
  \centering
  \includegraphics[width=\linewidth]{source/supp_inthewild.pdf}
  \caption{More results on in-the-wild scenarios.}
  \label{fig:Compare_more_itw}
\end{figure}

\begin{figure}[p]
  \centering
  \includegraphics[width=\linewidth]{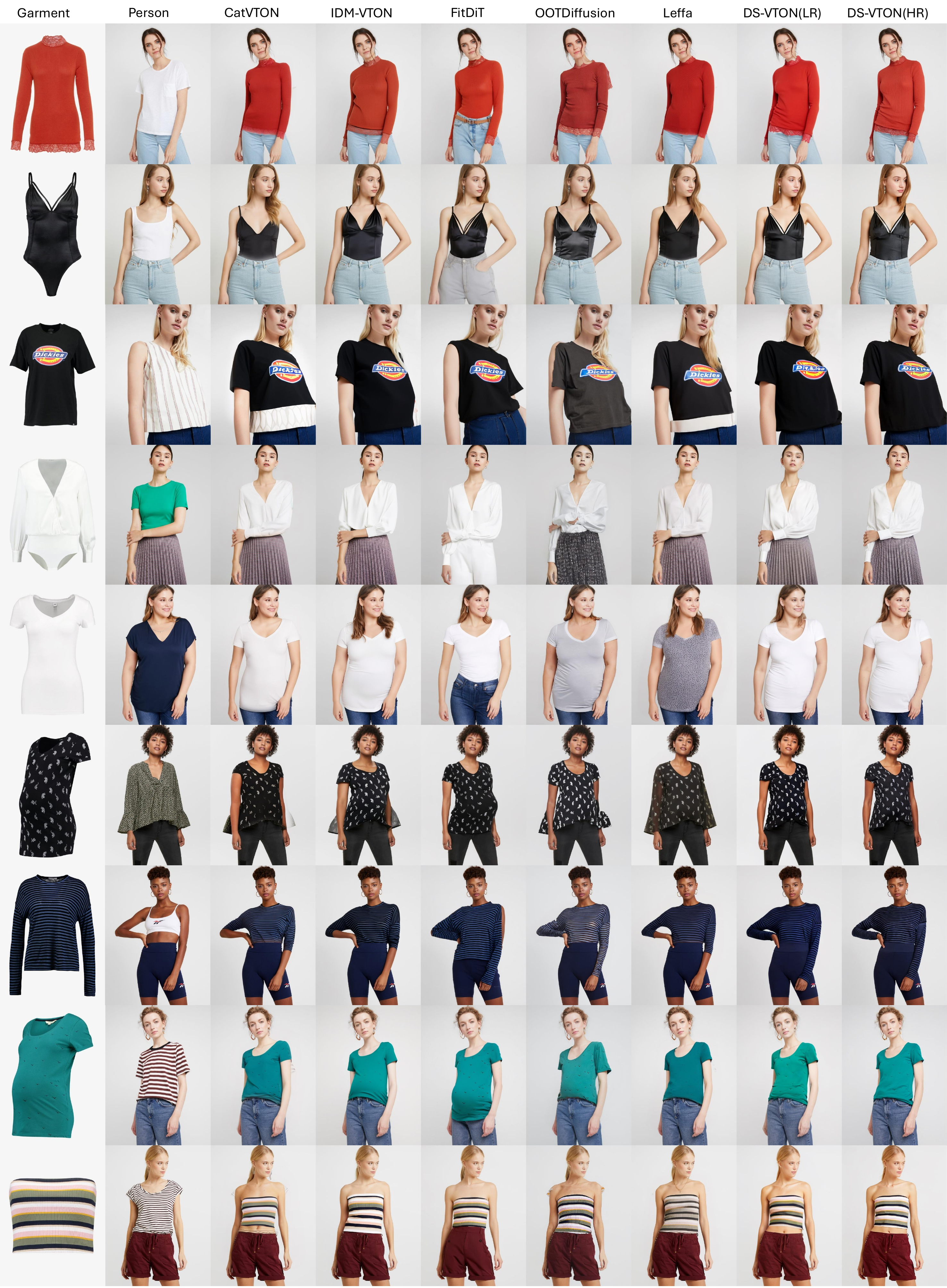}
  \caption{More qualitative comparison on the VITON-HD dataset. DS-VTON (LR) denotes the low-resolution output, and DS-VTON (HR) represents the final high-resolution result.}
  \label{fig:Compare_more}
\end{figure}

\begin{figure}[p]
  \centering
  \includegraphics[width=\linewidth]{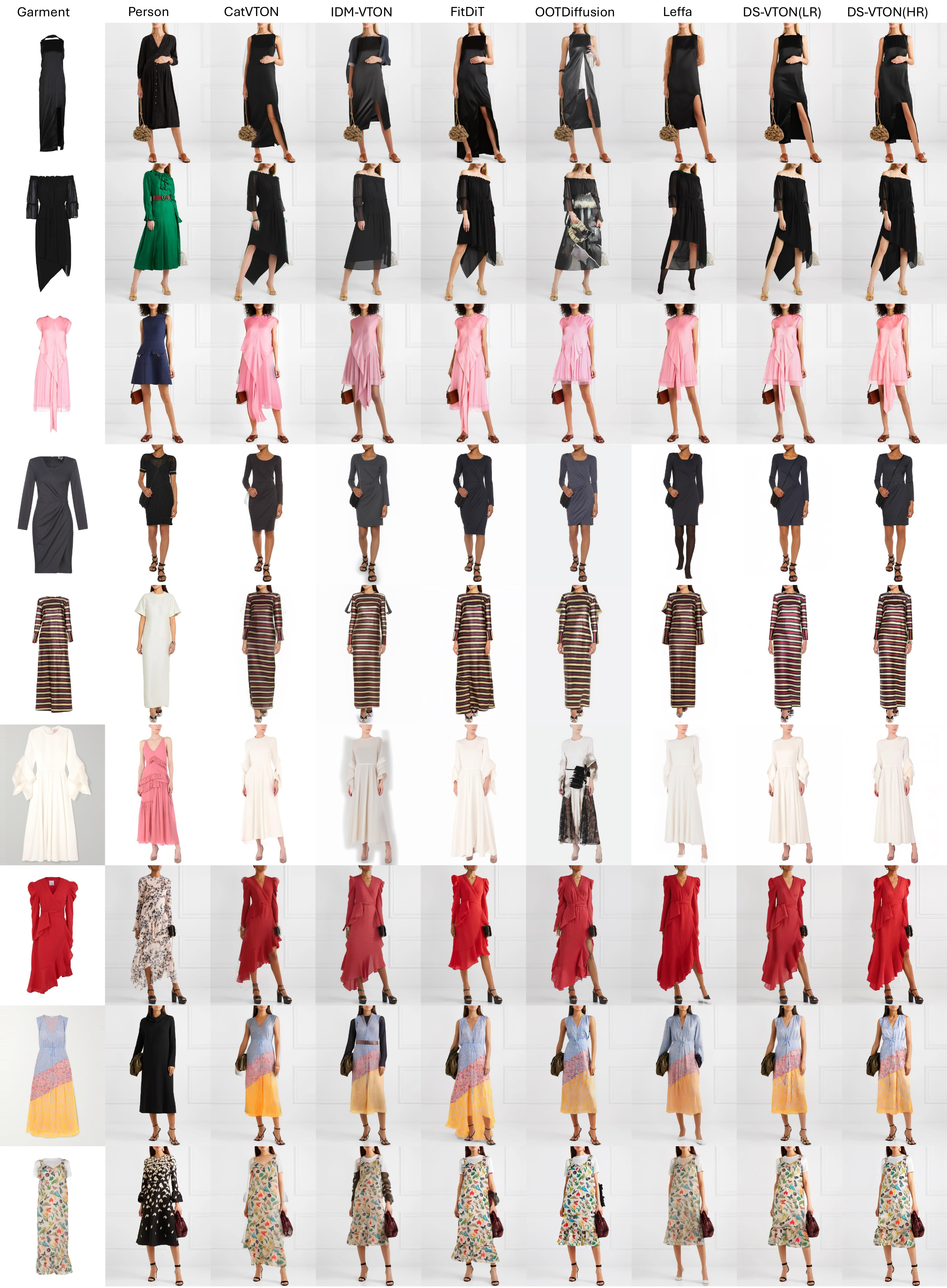}
  \caption{Qualitative comparison on the DressCode dataset (Dresses category). DS-VTON (LR) denotes the low-resolution output, and DS-VTON (HR) represents the final high-resolution result.}
  \label{fig:Compare_DressCode_Dress}
\end{figure}

\begin{figure}[p]
  \centering
  \includegraphics[width=\linewidth]{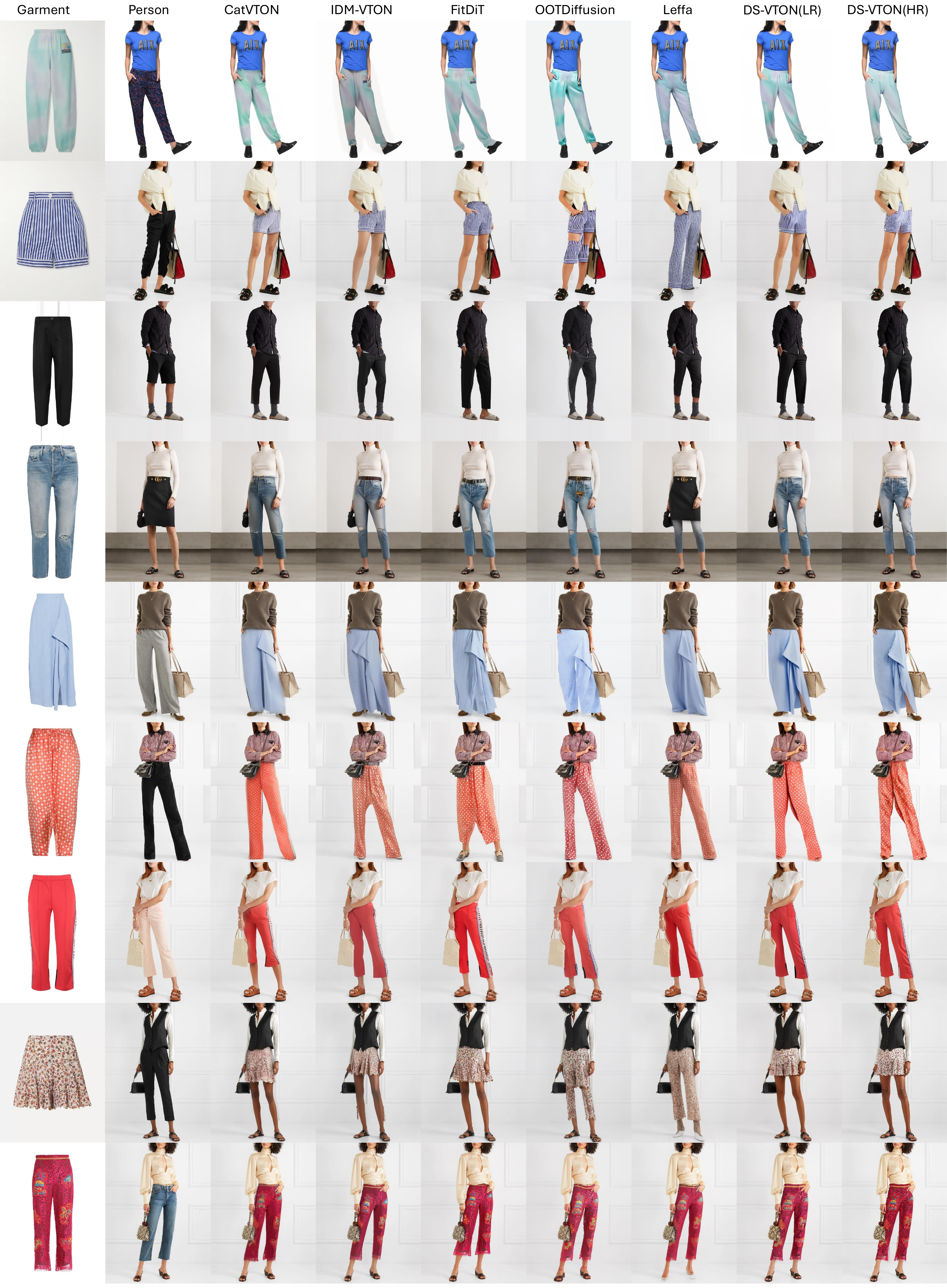}
  \caption{Qualitative comparison on the DressCode dataset (Lower category). DS-VTON (LR) denotes the low-resolution output, and DS-VTON (HR) represents the final high-resolution result.}
  \label{fig:Compare_DressCode_Lower}
\end{figure}

\begin{figure}[p]
  \centering
  \includegraphics[width=\linewidth]{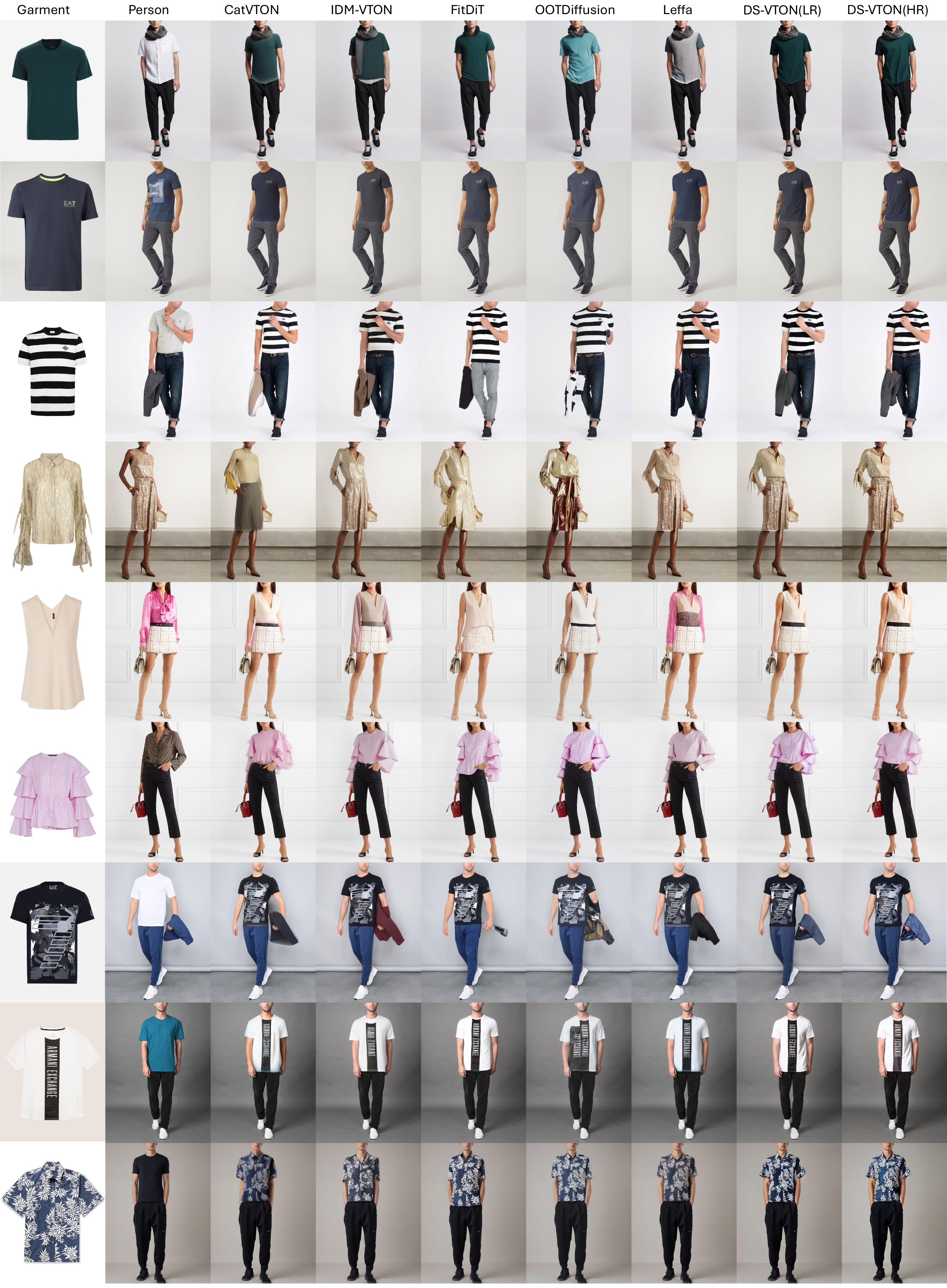}
  \caption{Qualitative comparison on the DressCode dataset (Upper category). DS-VTON (LR) denotes the low-resolution output, and DS-VTON (HR) represents the final high-resolution result.}
  \label{fig:Compare_DressCode_Upper}
\end{figure}

\end{document}